\documentclass[letterpaper]{article} 
\usepackage{aaai2026}  
\usepackage{times}  
\usepackage{helvet}  
\usepackage{courier}  
\usepackage[hyphens]{url}  
\usepackage{graphicx} 
\usepackage{amsmath}
\usepackage{amsthm}
\usepackage{amssymb}
 \usepackage{booktabs}
 \usepackage{float}

\usepackage{subcaption}
\usepackage{booktabs}
\usepackage{algorithm}
\usepackage{algorithmic}
\usepackage{placeins}
\usepackage{cleveref}
\usepackage{enumitem}
\usepackage{natbib}  
\usepackage{caption} 
\usepackage[caption=false]{subfig} 
\usepackage{newfloat}
\usepackage{listings}
\usepackage{bibentry} 

\DeclareCaptionStyle{ruled}{labelfont=normalfont,labelsep=colon,strut=off}
\floatstyle{ruled}
\newfloat{listing}{tb}{lst}{}
\floatname{listing}{Listing}

\title{Hierarchical Learning for Maze Navigation: Emergence of Mental Representations via Second-Order Learning}

\author {
    Shalima Binta Manir\textsuperscript{\rm 1},
    Tim Oates\textsuperscript{\rm 2}
}
\affiliations {
    \textsuperscript{\rm 1}Department of Computer Science, University of Maryland, Baltimore County\\
     \textsuperscript{\rm 2}Department of Computer Science, University of Maryland, Baltimore County\\
    smanir1@umbc.edu, oates@cs.umbc.edu
}

\begin{document}
\nocopyright

\maketitle

\begin{abstract}
Mental representation, characterized by structured internal models mirroring external environments, is fundamental to advanced cognition but remains challenging to investigate empirically. Existing theory hypothesizes that second-order learning -- learning mechanisms that adapt first-order learning (i.e., learning about the task/domain) -- promotes the emergence of such environment-cognition isomorphism. In this paper, we empirically validate this hypothesis by proposing a hierarchical architecture comprising a Graph Convolutional Network (GCN) as a first-order learner and an MLP controller as a second-order learner. The GCN directly maps node-level features to predictions of optimal navigation paths, while the MLP dynamically adapts the GCN's parameters when confronting structurally novel maze environments. We demonstrate that second-order learning is particularly effective when the cognitive system develops an internal mental map structurally isomorphic to the environment. Quantitative and qualitative results highlight significant performance improvements and robust generalization on unseen maze tasks, providing empirical support for the pivotal role of structured mental representations in maximizing the effectiveness of second-order learning.
\end{abstract}

\section{Introduction}
Understanding the computational mechanisms behind mental representation—the formation of internal cognitive structures aligned  with environmental structures—is crucial for advancing artificial cognitive systems. Historically, theories such as cognitive maps proposed by Tolman~\cite{tolman1948cognitive} have offered a framework to conceptualize these representations, emphasizing their role in enabling flexible navigation and efficient problem-solving. Arnold et al.~\cite{arnold2012second, arnold2015selection} proposed a theoretical model suggesting that such representational cognition emerges from evolutionary pressures favoring second-order learning, which involves mechanisms adapting first-order learning itself. In their framework, selection pressure on second-order learning results in selection pressure on isomorphism-based execution of first-order learning. Mental representation constitutes a component of this isomorphism.

The hypothesis that mental representation is a fundamental aspect of advanced cognition aligns with neuroscientific findings showing that the brain develops internal spatial representations isomorphic to environmental structure. For instance, studies on place and grid cells suggest that “place cells are the basic elements of a distributed noncentered map-like representation,” enabling organisms to perceive and remember their position in space~\cite{moser2008place}. These findings support the notion that internal cognitive structures (i.e., mental representations) emerge from systems that structurally mirror the external world.

Despite the theoretical appeal, there is limited direct empirical validation of how evolutionary selection for second-order learning induces a corresponding selection for first-order learning mechanisms that are implemented in a way isomorphic to the structure of the environment. Furthermore, most existing computational approaches rely explicitly on evolutionary algorithms, which, while powerful, can be computationally demanding and less interpretable in terms of learning dynamics.

In this paper, we provide empirical validation of Arnold et al.'s hypothesis using a novel hierarchical learning framework that combines Graph Convolutional Networks (GCNs) and second-order adaptive controllers implemented through Multi-Layer Perceptrons (MLPs). Specifically, the GCN serves as a first-order learner that predicts optimal paths in maze navigation tasks, while an MLP-based second-order learner dynamically adjusts the GCN's parameters in response to structurally novel maze environments. Importantly, we show that second-order learning significantly improves the adaptive capacity of the network, but its effectiveness is notably enhanced when an internal mental map—structurally isomorphic to the maze environment—emerges within the latent representations.

Our contributions are summarized as follows:
\begin{enumerate}
    \item We propose and implement a hierarchical second-order learning framework using GCNs and an MLP controller, demonstrating effective adaptation to novel maze structures without evolutionary optimization.
    \item We empirically confirm Arnold et al.'s \cite{arnold2012second} theoretical prediction that second-order learning facilitates the emergence of mental representations structurally aligned with the environment, quantitatively supported by high correlation metrics.
    \item We demonstrate substantial improvements in generalization and adaptation performance, highlighting the crucial role of structured, isomorphic internal representations in maximizing the efficiency and effectiveness of second-order learning.
\end{enumerate}

Our work thus bridges a critical gap between theoretical predictions and practical neural network implementations, providing concrete computational insights into the evolution and mechanisms underlying mental representation formation.

\section{Background and Related Works}
\textbf{Mental Representation and Cognitive Maps.} Mental representation, particularly through cognitive maps, provides a theoretical and empirical basis for understanding advanced cognition~\cite{tolman1948cognitive, banino2018vector, cueva2018emergence}. Cognitive maps enable agents to internally model environments, thus enhancing navigational flexibility and adaptability. Recent neural implementations, notably using graph neural networks, have shown promise in capturing spatial relationships and structural isomorphisms in tasks analogous to cognitive mapping~\cite{wu2020comprehensive, battaglia2018relational}.

\textbf{Second-Order Learning and Meta-Learning.} Second-order learning—learning processes that adapt the mechanisms of first-order learning—has roots in cognitive science as exemplified by Harlow’s classic learning-to-learn paradigm~\cite{harlow1949formation}. Computationally, meta-learning methods like Model-Agnostic Meta-Learning (MAML)~\cite{finn2017model, nichol2018first} operationalize second-order learning through gradient-based adaptations, achieving rapid generalization across novel tasks. Arnold et al.~\cite{arnold2012second, arnold2015selection} theoretically extended these ideas, suggesting that second-order learning exerts selection pressure favoring cognitive structures isomorphic with environmental features, thus driving the evolution of representational cognition.

\textbf{Adaptive and Hierarchical Graph Neural Networks.} Graph neural networks (GNNs), particularly Graph Convolutional Networks (GCNs)~\cite{kipf2017semi, velivckovic2017graph}, excel in representing graph-structured data and have been applied extensively in spatial navigation tasks. Recent work integrating meta-learning with GNNs, such as meta-adaptive GNNs~\cite{jiang2019meta, zugner2019adversarial}, demonstrates that adaptive hierarchical approaches significantly enhance generalization and adaptation capabilities. Yet, these methods have seldom examined how selection pressure on second-order learning could drive the emergence of isomorphism-based mechanisms in first-order learning of which mental representation is a central manifestation.

\textbf{Hypernetworks and Adaptive Weight Generation.} Hypernetworks provide a computational paradigm where one network dynamically generates the parameters for another~\cite{ha2016hypernetworks}. Initially inspired by evolutionary approaches like HyperNEAT~\cite{stanley2009hyperneat}, hypernetworks employ gradient-based training to achieve computationally efficient dynamic weight generation. They have been successfully applied in convolutional and recurrent architectures to dynamically adjust parameters for improved task-specific adaptation, such as in sequence modeling and neural machine translation~\cite{ha2016hypernetworks}. Our work extends this approach to hierarchical graph neural networks, leveraging similar dynamic parameter-generation mechanisms (via an MLP controller acting analogously to a hypernetwork) to enable second-order learning within graph-based maze navigation tasks.

\textbf{Our Positioning and Novelty.} Although the concepts of cognitive maps, second-order learning, adaptive GNNs, and hypernetworks have individually received significant attention, our approach uniquely integrates these strands. We specifically validate Arnold et al.'s \cite{arnold2012second} hypothesis regarding second-order learning's role in forming environment-cognition isomorphisms through a practical, non-evolutionary implementation. By doing so, we provide crucial empirical support for the theoretical claims in the literature, advancing the understanding of representational cognition and its underlying adaptive mechanisms in neural systems.
While previous studies have separately explored cognitive maps, second-order learning, and adaptive graph neural networks, there is limited direct empirical validation of second-order learning mechanisms explicitly facilitating environment-cognition isomorphism. This paper fills this gap by providing a clear computational model, without relying on evolutionary algorithms, that demonstrates how second-order learning effectively drives the emergence of structured, isomorphic representations.

\section{Methodology}
.Our framework investigates hierarchical learning through a combination of first-order and second-order learning mechanisms, specifically within a maze navigation task. Our primary aim is to empirically demonstrate how second-order learning facilitates the emergence of structured internal representations that align closely (are isomorphic) with the external maze environment.


We consider maze navigation tasks formulated as node classification problems, where a graph-based representation of the maze is given by \(G=(V,E)\), with nodes \(v \in V\) having associated features such as spatial coordinates, blockage information, and degrees of connectivity. The objective is to classify each node as being part of the shortest path from a start to a goal node.


First-order learning is implemented using a Graph Convolutional Network (GCN)\cite{kipf2017semi}, which directly maps node features to binary labels indicating path membership (on the path or not):
\begin{equation}
    L_1: X \rightarrow R
\end{equation}
where \(X\) is the stimulus (input features of nodes) and \(R\) is the response (binary classification indicating path nodes). Specifically, the GCN computes:
\begin{equation}
    Y = \text{GCN}(X, E; \theta)
\end{equation}
where \(X\) are node features, \(E\) is the edge index representing maze connectivity, and \(\theta\) are the GCN parameters. The GCN is trained using gradient descent to minimize a Binary Cross-Entropy (BCE) loss.



The second-order learner is implemented as an MLP-based controller, responsible for dynamically adapting the parameters of the GCN. Given a trained GCN and a novel maze scenario, the MLP controller learns to modify GCN parameters to achieve improved generalization and performance on structurally unknown tasks. Formally, this second-order adaptation can be expressed as:
\begin{equation}
    L_2: (Y, \theta) \rightarrow \theta'
\end{equation}
The adaptation mechanism takes two types of input: (1) the current outputs of the GCN $S$ and (2) the current parameter set \(\theta\). The MLP computes parameter updates as follows:
\begin{equation}
    \Delta\theta = \text{MLP}(\text{flatten}(Y), \text{flatten}(\theta))
\end{equation}

Adapted parameters \(\theta'\) are then derived as:

\begin{equation}
    \theta' = \theta + \Delta\theta
\end{equation}

The second-order learning itself is performed via gradient-based optimization of the MLP, where the MLP's parameters \(\phi\) are trained to minimize a BCE loss computed using adapted GCN predictions:

\begin{equation}
\begin{split}
\mathcal{L}_{\mathrm{MLP}}(\phi) = 
    -\frac{1}{M} \sum_{j=1}^{M} \sum_{i=1}^{N_j} \Big[
        y_{ij} \log(\hat{y}'_{ij}) \\
        + (1 - y_{ij}) \log(1 - \hat{y}'_{ij})
    \Big]
\end{split}
\end{equation}

where \(\hat{y}'_{ij}\) is the adapted prediction for node \(i\) in maze task \(j\), and \(M\) is the number of maze adaptation tasks used in training.

Hierarchical learning occurs through the sequential interaction of the GCN and the MLP controller. For more details, see Appendix.
\subsection{Emergence of Representational Isomorphism}

A central aspect of our methodology is verifying the emergence of internal cognitive representations isomorphic to maze geometry, which we quantify by computing correlations between pairwise distances in latent embeddings and actual spatial distances in the maze:

\begin{align}
    r_p &= \text{PearsonCorr}(D_{\text{latent}}, D_{\text{maze}}) \\
    r_s &= \text{SpearmanCorr}(D_{\text{latent}}, D_{\text{maze}})
\end{align}

High correlation values empirically confirm the representational alignment hypothesis. This can be viewed as representational alignment metric.


The Pearson product–moment coefficient measures linear metric fidelity and is most informative when the variables are approximately bivariate normal\cite{Pearson1896}.
Spearman’s rank‑order coefficient tests for a monotonic relationship, making it robust to non‑linear warping and outliers\cite{Spearman1904}.

\textbf{Interpretation.}  
In this setting, a \emph{high} Spearman coefficient (\(r_s\)) indicates that the pair‑wise ordering of distances in the latent space faithfully mirrors the ordering of true spatial distances—i.e., the topology of the maze is preserved. A \emph{high} Pearson coefficient (\(r_p\)) further shows that the mapping is approximately linear, meaning metric relations (relative magnitudes, ratios) are maintained as well.  
Conversely, if \(r_s\) is high but \(r_p\) is low, the embedding captures the correct qualitative layout (which state is nearer or farther) but distorts exact distances through non‑linear warping. If both coefficients are low, the latent geometry fails to align with the physical maze, falsifying the representational‑alignment hypothesis. Thus, the pattern of \(r_s\) and \(r_p\) values jointly diagnoses whether the model has learned a merely ordinal map, a metrically faithful map, or no coherent map at all.


Maze environments are generated algorithmically, creating controlled perturbations to assess adaptation robustness. Each node’s feature vector encodes coordinates and structural properties (blockages, degrees), systematically varying across experiments to test adaptation performance rigorously.
Our implementation includes delineated algorithmic procedures.
See Appendix for a detailed description of the algorithm.
The algorithmic setup captures the essence of the theoretical framework by showing that selection for second order learning yields an internal structure (mental representation) that mirrors the external environment.

To evaluate the broader utility of second-order learning beyond path classification, we extend our framework to \textbf{value-based maze tasks}, where the objective is to predict the \textbf{optimal cumulative reward} (value function) for each node in the maze.

We retain the hierarchical structure: a base GCN serves as the first-order learner, while an MLP controller implements second-order learning by adapting the GCN’s parameters in reward-perturbed environments.


The GCN is trained to regress value functions computed via dynamic programming (DP), where the reward for each node \((i, j)\) is defined as \(R(i,j) = i + j\), and the target value is recursively computed as:
\begin{equation}
V(i,j) = R(i,j) + \max(V(i+1,j), V(i,j+1))
\end{equation}
The GCN learns to approximate these DP-derived targets using MSE loss:
\begin{equation}
\mathcal{L}_{\text{GCN}} = \frac{1}{N}\sum_{i=1}^{N} (V_i - \hat{V}_i)^2
\end{equation}


The MLP controller adapts the GCN to environments with \textbf{altered reward structures}. Specifically, random multiplicative masks \(\in \{+1, -1\}\) simulate blocked or penalized regions in the maze. The MLP receives as input:

\begin{itemize}
    \item The predicted values from the base GCN,
    \item The current flattened parameters of the GCN.
\end{itemize}

It outputs adapted parameters \(\theta' = \theta + \Delta\theta\), computed via:

\begin{equation}
\Delta\theta = \text{MLP}(\text{flatten}(\hat{V}), \text{flatten}(\theta))
\end{equation}

The adapted GCN is evaluated against the DP value targets using MSE, and gradients are backpropagated through the MLP to optimize its parameters.
Our methodology operationalizes hierarchical learning and explicitly tests the theoretical claims of Arnold et al.~\cite{arnold2012second, arnold2015selection}, providing robust empirical validation of second-order learning as a driver for the emergence of structured mental representations. See appendix for more detail about evaluation metrics.

\section{Experiments and Results}
We empirically validate the efficacy of our hierarchical learning framework, composed of a first-order Graph Convolutional Network (GCN) and a second-order MLP Controller learning mechanisms, across a range of maze navigation and adaptation tasks. Our results demonstrate that second-order learning significantly improves generalization and internal representation structure under distributional shift.

\subsection{Experiment 1: Adaptation via MLP Controller}

In this experiment, we evaluate the ability of a second-order learner to adapt a GCN trained on unperturbed mazes to structurally altered environments.

\textbf{Setup:} We train a base GCN to predict shortest-path node labels on a \(10 \times 10\) grid maze, where each node is connected to its immediate neighbors (up, down, left, and right), except at the grid boundaries. The starting position is always at the top-left corner, denoted as \((0, 0)\) and the goal is fixed at the bottom-right corner, denoted as\((9, 9)\). The test mazes are generated by randomly removing the edges (using the block probability \(p \in \{0.1, 0.2, 0.3, 0.4, 0.5\}\)), forming a distribution shift. We take three new mazes to test. An MLP controller receives the GCN’s output and current weights, 
and predicts updated parameters for adaptation.

\textbf{Result:} We compare the performance of a base GCN (Unadapted GCN) and our proposed MLP-adaptive GCN (Adapted GCN) in unseen mazes. The Adapted GCN achieves significantly higher accuracy, averaging 93.6\% in unseen environments, compared to just 61\% for the Unadapted GCN.
Also, we compare the binary cross-entropy (BCE) loss between the Unadapted GCN and the Adapted GCN on structurally altered mazes. The unadapted model, trained solely on clean 10$\times$10 grid mazes, exhibits significantly higher loss on perturbed test mazes (BCE $\approx$ 1.08), indicating poor generalization under distribution shift. In contrast, the Adapted GCN, which leverages an MLP controller to modify the GCN’s weights based on its internal state and outputs, achieves a substantially lower loss (BCE $\approx$ 0.42).

\subsection{Latent Representation Visualization:} To examine the internal representations learned by the model, we apply t-SNE to the node embeddings produced by the first layer of the GCN.
\paragraph{Why t\textendash SNE?}
To probe the geometry of the learned representation, we require a dimensionality reduction method that (i) faithfully preserves the structure of the local neighborhood, (ii) highlights objectively discoverable clusters and (iii) produces visually interpretable layouts.  
t\textendash SNE satisfies these criteria by minimising the Kullback–Leibler divergence between pairwise similarities in the high‑ and low‑dimensional spaces, thereby maintaining local topology while exaggerating larger‑scale separations\cite{van2008visualizing}.  
This makes emergent “mental maps’’ encoding both spatial proximity and functional equivalence—readily visible.  
Alternative techniques such as PCA inadequately capture non‑linear relations \cite{abdi2010principal}. 
Hence, t\textendash SNE offers the clearest qualitative evidence that the GCN’s latent space is structured in a way that is isomorphic to the external maze.
\paragraph{Latent Representation Learning and Functional Clustering}
We first ask whether the GCN internalises a notion of \emph{path–relevance}.
After training the network to label every node in a \(10\times10\) maze as either on– or off–the shortest path from the start \((0,0)\) to the goal \((9,9)\), we extract the latent vectors produced by the first GCN layer.\footnote{The dimensionality \(d\) of these vectors equals the \texttt{hidden\_dim} hyper‑parameter in the model definition; all results reported here are agnostic to its specific value.}
We then project the \(d\)-dimensional embeddings onto two dimensions with t‑SNE for visual inspection.

\begin{figure*}[t]
    \centering
    \begin{subfigure}[b]{0.32\textwidth}
        \centering
        \includegraphics[width=\textwidth]{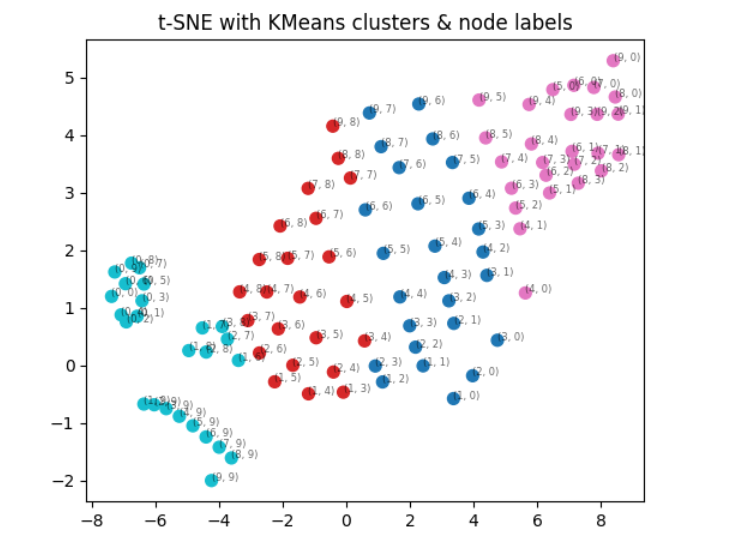}
        \caption{t-SNE plot showing optimal path nodes (red) versus other nodes (gray). Separation indicates meaningful internal abstraction of path information.}
        \label{fig4}
    \end{subfigure}
    \hfill
    \begin{subfigure}[b]{0.32\textwidth}
        \centering
        \includegraphics[width=\textwidth]{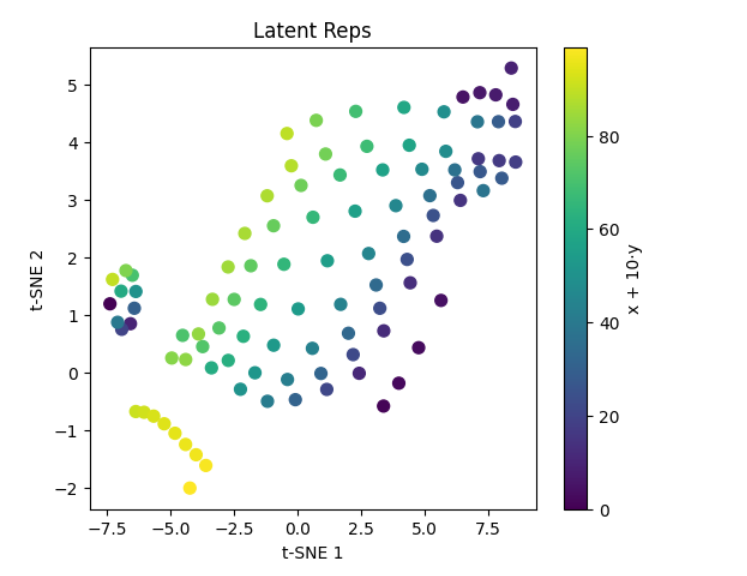}
        \caption{Latent space colored by spatial coordinate index ($x + 10y$). Clear gradients suggest spatial structure is preserved in the learned representations.}
        \label{fig5}
    \end{subfigure}
    \hfill
    \begin{subfigure}[b]{0.32\textwidth}
        \centering
        \includegraphics[width=\textwidth]{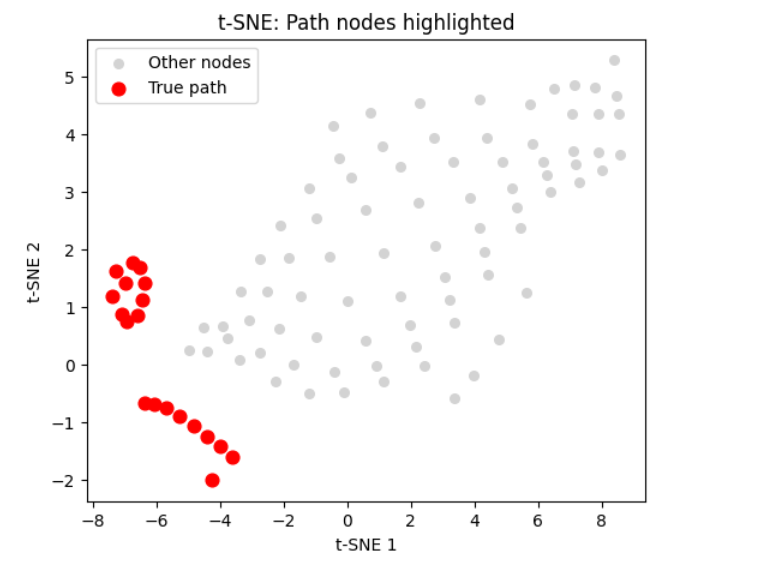}
        \caption{t-SNE plot showing optimal path nodes (red) versus other nodes (gray). Separation indicates meaningful internal abstraction of path information.}
        \label{fig6}
    \end{subfigure}
    \caption{Alignment between representations and the environment.}
    \label{fig:all_three}
\end{figure*}

Clustering the same embeddings with KMeans \((k=4\), selected via the elbow criterion). \Cref{fig4} shows a 2D t-SNE projection of node embeddings of experiment 1, where colors indicate KMeans cluster assignments and node labels reflect original maze coordinates. The clusters are clearly formed and spatially organized, indicating that the GCN develops distinct internal representations for different node roles.

\paragraph{Global Structure of the Latent Space}
\Cref{fig4} overlays the t‑SNE projection with both the KMeans assignments (colour) and the original maze coordinates (text label).
The clusters are not arbitrary clouds: they form an orderly “patchwork’’ that respects the grid topology, implying that the network has learnt to encode \emph{where} a node is \emph{and} \emph{what it does}.
In particular, neighbouring coordinates tend to lie close together in the embedding, while nodes serving different functional roles occupy distinct regions.

\paragraph{Structural Consistency in Latent Representations}
To assess how well the latent space from the first GCN layer, trained on the original maze, captures the structure of the external environment, we visualize the embeddings by coloring each point according to its linearized spatial index \(x + 10y\) \Cref{fig5}.
The almost monotonic colour gradient demonstrates a striking isomorphism: Euclidean distance in the embedding correlates with Manhattan distance in the maze. The visualization reveals that nodes with similar spatial positions in the maze tend to cluster together in the latent space, indicating that the GCN captures meaningful topological structure(connectivity pattern).
Thus, without explicit supervision on coordinates, the GCN reconstructs a map‑like structure internally.

\paragraph{Distinct Latent Region for the Optimal Path}
\Cref{fig6} highlights the latent positions of the optimal path nodes (in red) compared to all other nodes (in gray). Their tight clustering indicates the GCN distinctly encodes path-relevant nodes, supporting the idea of internalized path abstraction or "mental mapping."  All nodes on the path are assigned to \textbf{Cluster 3}, suggesting that GCN successfully learns to embed structurally or functionally similar nodes in proximity.

\paragraph{t‑SNE and Isomorphism}
To assess whether the model's internal representations preserve the structure of the external environment, we use t-SNE to project high-dimensional latent node embeddings into two dimensions. This visualization technique helps reveal isomorphism by showing that nodes which are spatially close in the maze often appear close in the latent space as well. The clear clustering of optimal-path nodes and the smooth gradient of spatial coordinates in the t-SNE plots indicate that the model develops a structured internal "mental map" of the environment, supporting the emergence of environment-cognition isomorphism. 

\subsection{Spatial Coherence of the Latent Space}
t‑SNE projections condense the high–dimensional node embeddings into two
dimensions and immediately expose coherent spatial clusters: nodes that are
adjacent on the physical grid appear side‑by‑side in the plot, while
path‑relevant nodes occupy a compact, separate region.
To quantify this visual intuition we correlate the pair‑wise Euclidean
distances in latent space with the true Euclidean distances on the maze. The distance matrices are computed from the hidden vectors output by the \emph{first} GCN layer (after the ReLU activation), i.e.\ the model’s internal node embeddings prior to the classification head.
In experiment 1, both the Pearson coefficient \((r_{p}=0.9042)\) and the Spearman coefficient
\((r_{s}=0.9085)\) lie close to their theoretical maximum of~1, implying that
closeness in the embedding is almost perfectly aligned—both linearly and
monotonically—with closeness in the real environment.  
These high correlations provide clear evidence that the model has constructed
an internal \emph{mental map}: its representation preserves the maze’s spatial
relationships rather than encoding nodes arbitrarily, thereby demonstrating
environment–cognition isomorphism.



\subsection{Controller training dynamics.}
\begin{figure}[t]
\centering
\includegraphics[width=0.9\columnwidth]{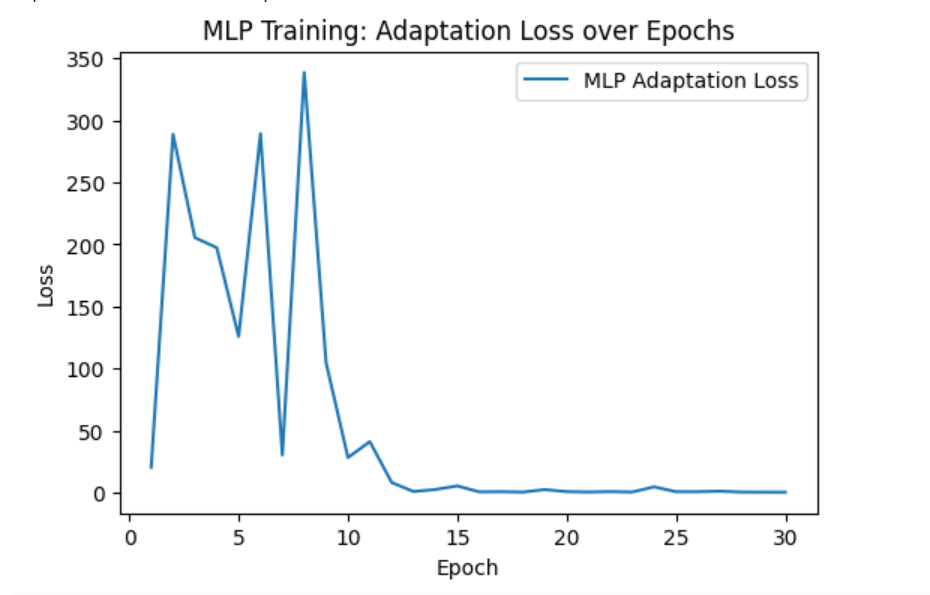} 
\caption{Training loss of the MLP Controller across epochs.}
\label{fig:exp1-ctrl}
\end{figure}

The MLP’s loss decreases smoothly across epochs
(\Cref{fig:exp1-ctrl}), confirming stable second‑order optimisation.
Across all blocked mazes, the adapted GCN achieves lower MSE and higher policy accuracy than the unadapted baseline. This confirms that the learned adaptation mechanism produces functionally useful changes, rather than merely synthetic parameter updates. Furthermore, the controller generalizes across tasks, block rates, and maze layouts indicating a non-trivial, transferable inductive bias.

\subsection{Experiment 2: Enriched Latent Input for MLP-based GCN Adaptation}

In this experiment we want to investigate whether deeper internal representations improve MLP-based adaptation. The MLP Controller is fed a concatenation of: Flattened hidden activations (GCN layer 1), Flattened output logits, Flattened GCN parameters.
This results in an augmented input vector of dimensionality:
\begin{equation}
\begin{split}
\text{Input Dim} \;=\;& (\text{hidden\_dim} \times \text{num\_nodes})\\
                     &+ (\text{output\_dim} \times \text{num\_nodes})\\
                     &+ \text{num\_params}
\end{split}
\end{equation}

\textbf{Results:} The MLP adaptor supplied with hidden‐layer activations attains a mean accuracy of 93.6\% on the three unseen mazes, while the average binary‑cross‑entropy loss falls to 0.2335. Although the enriched input does not raise accuracy beyond that of the baseline adaptor, it supports more efficient optimisation, as evidenced by the markedly lower loss.

\subsection{Experiment 3: Generalization Across Maze Sizes}

In this experiment, we examine the ability of a GCN trained solely on 8×8 mazes to generalize spatial representations to larger maze sizes. The training process used coordinate features (row, column) and optimized the latent space to preserve pairwise grid distances. We then adapted the model using an MLP trained on blocked versions of the same\((8,8)\) maze, using block probabilities between 0.1 and 0.3.
The Pearson correlation between Euclidean distances in the GCN's latent space and the true grid distances was used as a proxy for spatial structure preservation. This correlation was computed after the first GCN layer, following the ReLU activation, providing a direct measure of how well the learned representation captures the underlying maze geometry. We perform two types of tests.
\begin{itemize}
    \item \textbf{Empirical study 1:} The GCN is trained on 8×8 mazes and evaluated directly on larger, unblocked mazes without any additional size-based training. It achieves:
    \begin{itemize}
        \item 10×10 maze: Pearson correlation = \textbf{0.9462}
        \item 12×12 maze: Pearson correlation = \textbf{0.9559}
        \item 14×14 maze: Pearson correlation = \textbf{0.9623}
    \end{itemize}
    
    \item \textbf{Empirical study 2:} The model is trained on both 8×8 and 10×10 mazes to incorporate more size variation. When evaluated on larger sizes:
    \begin{itemize}
        \item 12×12 maze: Pearson correlation = \textbf{0.9570}
        \item 14×14 maze: Pearson correlation = \textbf{0.9631}
    \end{itemize}
\end{itemize}
These results demonstrate that the GCN embeddings generalize remarkably well to larger mazes, even when trained on smaller ones. The strong correlation values indicate that the learned latent space preserves the spatial geometry of the environment. Notably, adding training diversity (as in Experiment 2) offers a slight but consistent improvement in generalization.
The trend of the results confirm that spatial alignment in the latent space is robust and improves slightly as test size increases. This suggests that the GCN learns general spatial principles rather than overfitting to a specific maze size. These results reinforce the idea that GCNs inherently develop spatially meaningful representations that transfer across scale. When paired with adaptation strategies like the MLP controller, this architecture is capable of supporting second-order learning — adjusting to environmental variation while preserving core structural understanding. This supports the broader view of latent representations as emergent, isomorphic cognitive maps.

\subsection{Experiment 4: Representation Collapse in Non-Isomorphic Mazes}

We examine the role of isomorphic feature-geometry mapping in enabling internal structure.
We replace spatially meaningful node features with i.i.d.\ Gaussian noise while maintaining the same graph structure. Although the mazes are visually and structurally identical (same connectivity), they no longer provide spatial signals through features. This isolates the impact of representational continuity on GCN performance. 

\textbf{Performance comparison: isomorphic vs. non-Isomorphic}

We evaluate the robustness of the Adapted GCN under structural perturbations by comparing its performance on isomorphic and non-isomorphic mazes. The model achieves an average test accuracy of approximately 91\% on isomorphic mazes, which drops to around 76\% on non-isomorphic mazes. Here, we take three isomorphic and three non-isomorphic mazes for evaluation.
We analyze the Adapted GCN's loss on isomorphic and non-isomorphic mazes. The model exhibits a low loss of approximately 0.2 on isomorphic mazes, which increases sharply to about 1.85 on non-isomorphic mazes. This substantial rise in loss indicates that loss increases significantly in the absence of geometric feature cues in non-isomorphic maze environment.

\begin{figure*}[t]
  \centering
  
  \begin{subfigure}[b]{0.31\textwidth}
    \centering
    \includegraphics[width=\linewidth]{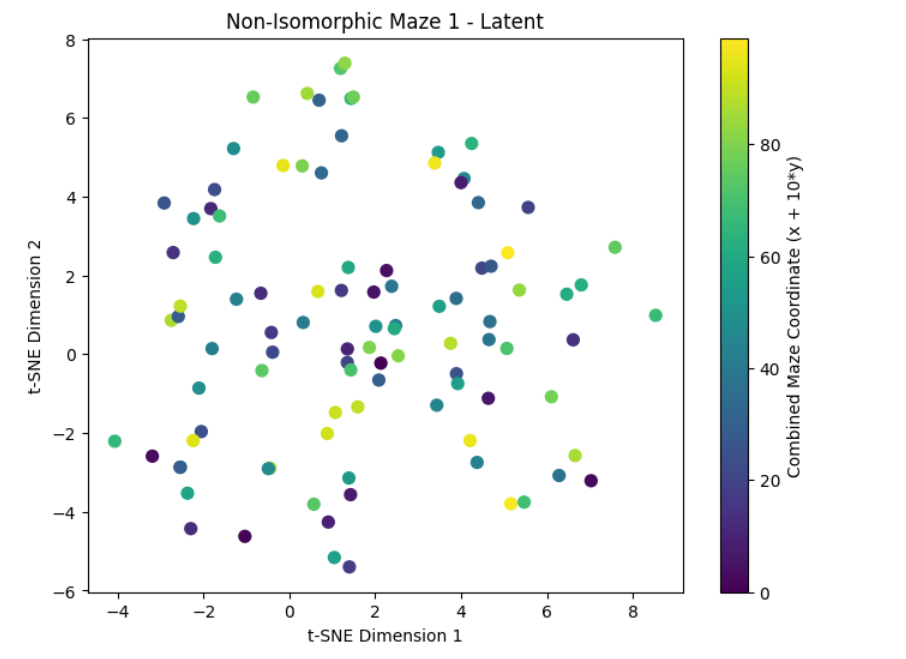}
    \caption{t-SNE plot of Non-Isomorphic Maze~1 (no spatial alignment).}
    \label{fig12}
  \end{subfigure}
  \hfill
  \begin{subfigure}[b]{0.31\textwidth}
    \centering
    \includegraphics[width=\linewidth]{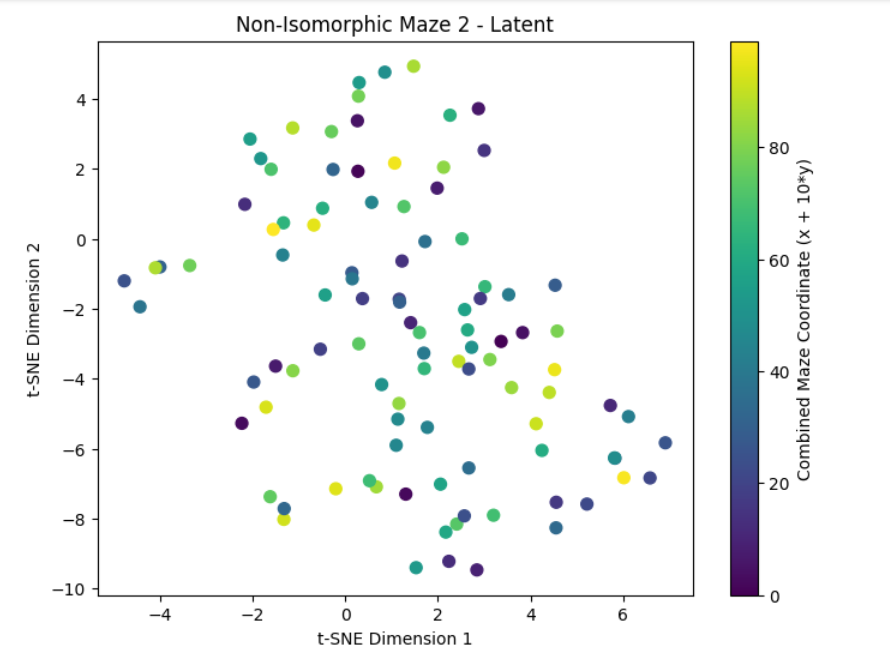}
    \caption{t-SNE plot of Non-Isomorphic Maze~2 (no spatial alignment).}
    \label{fig13}
  \end{subfigure}
  \hfill
  \begin{subfigure}[b]{0.31\textwidth}
    \centering
    \includegraphics[width=\linewidth]{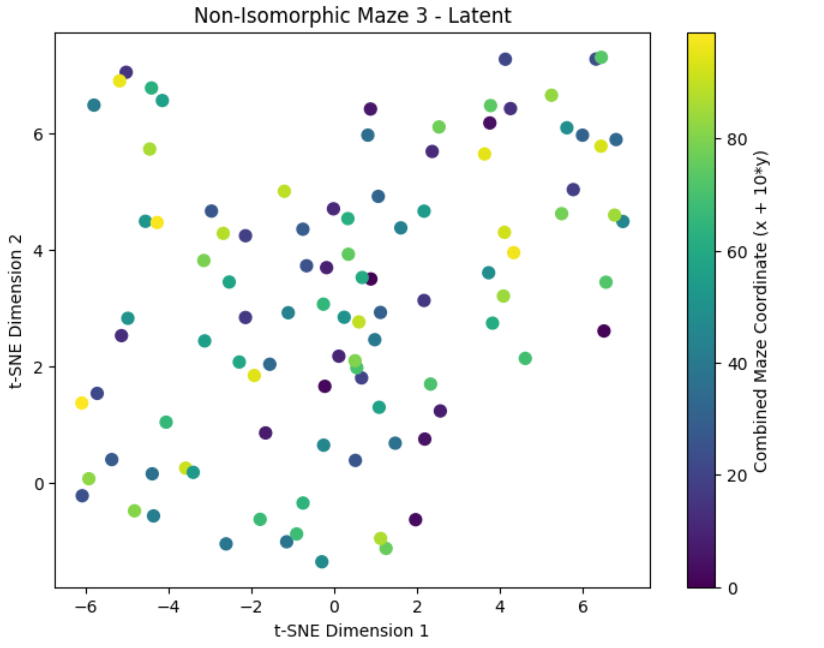}
    \caption{t-SNE plot of Non-Isomorphic Maze~3 (no spatial alignment).}
    \label{fig14}
  \end{subfigure}

  \caption{t-SNE visualizations for three non-isomorphic mazes, all showing a lack of spatial alignment in the learned latent space.}
  \label{fig:noniso_tsne_row}
\end{figure*}

To assess spatial alignment in the latent space, we compute the average pearson correlation between isomorphic and non-isomorphic three unseen mazes. The Adapted GCN achieves a high average correlation of approximately 0.86 on isomorphic mazes, indicating strong spatial structure preservation and also providing quantitative evidence for an emergent geometry–latent isomorphism. In contrast, this correlation drops drastically to around 0.05 on non-isomorphic mazes, reflecting a collapse in spatial consistency.
To visualize the disruption in learned representations, t-SNE plots are shown for three non-isomorphic mazes in \Cref{fig:noniso_tsne_row}. In contrast to isomorphic environments, where the latent space reflected spatial structure, these maps appear randomly distributed.

These findings confirm that meaningful feature-geometry correspondence is critical for effective adaptation and representation formation.

\textbf{Quantitative Results} Each test maze was evaluated in both adapted and unadapted forms:

\begin{itemize}
    \item \textbf{Maze 1:} Pearson corr = -0.0248 \\
          Adapted → Accuracy = 0.78, Loss = 1.8142 \\
          Unadapted → Accuracy = 0.72, Loss = 0.5588
    \item \textbf{Maze 2:} Pearson corr = -0.0580 \\
          Adapted → Accuracy = 0.70, Loss = 3.0707 \\
          Unadapted → Accuracy = 0.55, Loss = 0.6785
    \item \textbf{Maze 3:} Pearson corr = -0.0579 \\
          Adapted → Accuracy = 0.78, Loss = 0.7160 \\
          Unadapted → Accuracy = 0.54, Loss = 0.7216
\end{itemize}

Performance deteriorates on non-isomorphic tasks, demonstrating that the MLP controller leverages the environment–cognition isomorphism—an internal mapping to the environment—rather than relying on memorization.

\subsection{Experiment 5: Second-Order Learning for Value Adaptation}

The objective of this experiment is to evaluate second-order learning in a value prediction task using dynamic reward environments. The starting position is at (0, 0), and the goal is in the bottom right corner.

\textbf{Procedure:} The GCN is trained to predict cumulative rewards from each node to the goal (supervised by dynamic programming). The MLP Controller adapts the GCN under reward perturbations.
\textbf{Quantitative Results (unseen blocked mazes):}
\begin{itemize}
    \item \textbf{Unadapted GCN:} MSE = 297.47, MAE = 14.73, \(R^2 = -2.16\)
    \item \textbf{Adapted GCN:} MSE = 33.50, MAE = 4.99, \(R^2 = 0.64\)
    \item \textbf{Policy accuracy} rises from 25.0\% with the unadapted GCN to 66.7\% when the MLP-adapted GCN is used.
.
\end{itemize}

The second-order adaptation significantly improves both value prediction and policy alignment with optimal trajectories.
All the results provides compelling evidence supporting second-order learning's efficacy and potential, confirming its role in enabling rapid adaptation and robust representation.

Our results offer empirical validation of second-order learning dynamics. As posited in recent work, selection pressure at one level (here, the MLP adapting the GCN) leads to structure one level below (in the GCN’s representations).

We propose that the isomorphic alignment between a learner’s first-order representations and the environmental structure underpins and supports the emergence of second-order learning capabilities. In other words, the system’s capacity to learn how to learn (second order) relies fundamentally on the presence of isomorphic mappings formed during first-order learning.

\section{Conclusion}

We present a hierarchical maze-navigation architecture in which a first-order Graph Convolutional Network (GCN) serves as a stimulus–response learner, while a second-order MLP controller meta-learns to adaptively retune the GCN in response to structural changes in the environment. Empirical results across a range of conditions—including blocked paths, enlarged layouts, randomized feature representations, and value-function regression—demonstrate that the controller substantially improves performance: reducing error by up to an order of magnitude, shifting \(R^{2}\) from negative to positive, and more than doubling policy accuracy. Notably, adaptation fails when the altered maze is no longer isomorphic to the latent graph representation, highlighting the critical role of structural alignment in effective generalization.
These findings suggest that second-order learning mechanisms can induce internal cognitive maps that support flexible and robust reasoning in dynamic, graph-structured environments.

Broader impacts of this work include insights into forces that drive isomorphism between mental representations and environments in humans.  Future work will address limitations, such as expanding to more domains, more tasks, and more diverse learning algorithms of both the first and second order variety. 

\bibliography{sol_paper}


\appendix
\section*{Appendix}
\FloatBarrier  
\subsection{Algorithmic Description}

Below, we detail the main algorithms in pseudocode.

\begin{algorithm}[!h]
\footnotesize                 
\caption{Create Maze Graph}
\begin{algorithmic}[1]
\STATE \textbf{Input:} Maze size $n$, block probability $p$
\STATE $G \gets$ 2D grid graph of size $n \times n$
\STATE $\text{original\_edges} \gets E(G)$
\STATE $\text{original\_degrees} \gets \{ \deg(v) \mid v \in V(G) \}$
\STATE $\text{edges} \gets \text{List}(E(G))$
\STATE Shuffle $\text{edges}$
\STATE $\text{num\_blockages} \gets \lfloor p \cdot |\text{edges}| \rfloor$
\STATE $\text{removed\_edges} \gets \varnothing$
\FOR{edge in first $\text{num\_blockages}$ edges}
  \STATE Remove \textbf{edge} from $G$
  \STATE Append \textbf{edge} to $\text{removed\_edges}$
\ENDFOR
\STATE \textbf{For each} node $v$ in $G$:
\STATE \hspace*{1.5em}Compute feature vector $f(v) = [x(v), y(v),$
\STATE \hspace*{1.5em}$\text{\# blockages at } v,\ \text{original degree},\ \text{current degree}]$
\STATE $\text{node\_features} \gets \{v \mapsto f(v)\}$
\STATE \textbf{Output:} $G$, $\text{edge\_index}$, $\text{node\_features}$,
       $\text{original\_edges}$, $\text{removed\_edges}$, start, goal
\end{algorithmic}
\end{algorithm}

\begin{algorithm}[!h]
\footnotesize                    
\caption{Train GCN on Original Maze}
\begin{algorithmic}[1]
\STATE \textbf{Input:} Maze graph $G$, node features, edge index, labels derived from the maze shortest path.
\STATE Initialize GCN with parameters $\theta$.

\FOR{$\text{epoch} = 1$ to $T$}
  \STATE Forward propagate:
  \STATE \hspace*{1.5em}$\hat{y} \gets \text{GCN}(\text{node\_features}, \text{edge\_index};\ \theta)$
  \STATE Compute loss:
  \STATE \hspace*{1.5em}$\mathcal{L} \gets \text{BCE}(\hat{y}, y)$
  \STATE Backpropagate and update parameters:
  \STATE \hspace*{1.5em}$\theta \leftarrow \theta - \eta \nabla_\theta \mathcal{L}$
\ENDFOR

\STATE \text{Visualize latent representations using t-SNE.}
\STATE \text{Compute distance correlations between latent space and maze coordinates.}
\STATE \textbf{Output:} Trained GCN.
\end{algorithmic}
\end{algorithm}

\begin{algorithm}[!h]
\footnotesize                    
\caption{Train MLP Controller for GCN Adaptation}
\begin{algorithmic}[1]
\STATE \textbf{Input:} Set of blocked maze tasks $\{G_i\}$, pretrained GCN with parameters $\theta$, number of epochs $T_2$.

\FOR{$\text{epoch} = 1$ to $T_2$}
  \FOR{each maze task $i$}
    \STATE Compute GCN output on task $i$:
    \STATE \hspace*{1.5em}$\hat{y}_i \gets \text{GCN}(\text{features}_i, \text{edge\_index}_i;\ \theta)$
    \STATE Flatten $\hat{y}_i$ and current GCN weights $\theta$ into a single vector.
    \STATE MLP Controller outputs:
    \STATE \hspace*{1.5em}$\Delta \theta_i \gets \text{MLP}(\hat{y}_i, \theta)$
    \STATE Form adapted parameters: $\theta_i' \gets \Delta \theta_i$
    \STATE Compute adapted output:
    \STATE \hspace*{1.5em}$\hat{y}_i' \gets \text{GCN}(\text{features}_i, \text{edge\_index}_i;\ \theta_i')$
    \STATE Compute loss: $\mathcal{L}_i \gets \text{BCE}(\hat{y}_i', y_i)$
  \ENDFOR
  \STATE Compute total loss:
  \STATE \hspace*{1.5em}$\mathcal{L}_{\text{total}} = \frac{1}{N}\sum_i \mathcal{L}_i$
  \STATE Backpropagate and update MLP parameters.
\ENDFOR

\STATE \textbf{Output:} Trained MLP Controller.
\end{algorithmic}
\end{algorithm}

\begin{algorithm}[!h]
\footnotesize                    
\caption{Evaluation and Distance Correlation Computation}
\begin{algorithmic}[1]
\STATE \textbf{Input:} Unseen maze task $G$, features, edge index.
\STATE Obtain latent representations:
\STATE \hspace*{1.5em}$Z \gets \text{ReLU}(\text{GCN.conv1}(\text{node\_features}, \text{edge\_index}))$
\STATE Compute pairwise distances $D_Z$ in the latent space.
\STATE Extract maze coordinates $M$ for each node and compute pairwise distances $D_M$.
\STATE Flatten the upper triangles of $D_Z$ and $D_M$ to vectors $z$ and $m$.
\STATE Compute Pearson correlation $r_p$ and Spearman correlation $r_s$ between $z$ and $m$.
\STATE \textbf{Output:} $r_p$, $r_s$ (high values indicate strong preservation of maze geometry in $Z$).
\end{algorithmic}
\end{algorithm}

\subsection{Evaluation metrics}
\paragraph{Path–prediction experiment (shortest–path classification)}
\begin{itemize}[leftmargin=*]
    \item \textbf{Binary‑cross‑entropy (BCE) loss \& classification accuracy:} measured on {unseen} blocked mazes to quantify generalisation of the path/non‑path discriminator.
    \item \textbf{Representational isomorphism:} Pearson and Spearman correlations between (i) pairwise Euclidean distances in the first‑layer latent space and (ii) pairwise Manhattan distances in the maze layout.  High positive values indicate that the internal geometry mirrors the external environment.
\end{itemize}

\paragraph{Value‑adaptation experiment (dynamic reward map).}
\begin{itemize}[leftmargin=*]
    \item \textbf{Mean‑squared error (MSE)} and \textbf{mean‑absolute error (MAE)} between predicted and ground‑truth state‑value functions.
    \item \textbf{Coefficient of determination ($R^{2}$):} proportion of variance in the true values explained by the model.
    \item \textbf{Policy accuracy:} fraction of nodes at which the adapted model’s recommended action matches the optimal action obtained via dynamic programming.
\end{itemize}
These metrics capture both regression accuracy and policy-level behavior, providing a comprehensive evaluation of second-order learning performance in dynamic reward environments.

\subsection{Hierarchical Learning Procedure}
\begin{enumerate}
    \item \textbf{First-order training:} Train the base GCN on an initial set of unperturbed maze navigation tasks to convergence, establishing baseline parameters \(\theta\).
    \item \textbf{Second-order training (adaptation phase):} Given a set of structurally perturbed maze environments (e.g., mazes with randomly blocked paths), the MLP controller learns to dynamically adapt the GCN parameters. Adaptation is performed in a differentiable manner, allowing backpropagation from the GCN outputs through the adapted parameters to the MLP controller.
\end{enumerate}
\appendix
\makeatletter
\@ifundefined{isChecklistMainFile}{
  \newif\ifreproStandalone
  \reproStandalonetrue
}{
  \newif\ifreproStandalone
  \reproStandalonefalse
}
\makeatother

\ifreproStandalone
\documentclass[letterpaper]{article}
\usepackage[submission]{aaai2026}
\setlength{\pdfpagewidth}{8.5in}
\setlength{\pdfpageheight}{11in}
\usepackage{times}
\usepackage{helvet}
\usepackage{courier}
\usepackage{xcolor}
\frenchspacing

\begin{document}
\fi
\setlength{\leftmargini}{20pt}
\makeatletter\def\@listi{\leftmargin\leftmargini \topsep .5em \parsep .5em \itemsep .5em}
\def\@listii{\leftmargin\leftmarginii \labelwidth\leftmarginii \advance\labelwidth-\labelsep \topsep .4em \parsep .4em \itemsep .4em}
\def\@listiii{\leftmargin\leftmarginiii \labelwidth\leftmarginiii \advance\labelwidth-\labelsep \topsep .4em \parsep .4em \itemsep .4em}\makeatother

\setcounter{secnumdepth}{0}
\renewcommand\thesubsection{\arabic{subsection}}
\renewcommand\labelenumi{\thesubsection.\arabic{enumi}}

\newcounter{checksubsection}
\newcounter{checkitem}[checksubsection]

\newcommand{\checksubsection}[1]{%
  \refstepcounter{checksubsection}%
  \paragraph{\arabic{checksubsection}. #1}%
  \setcounter{checkitem}{0}%
}

\newcommand{\checkitem}{%
  \refstepcounter{checkitem}%
  \item[\arabic{checksubsection}.\arabic{checkitem}.]%
}
\newcommand{\question}[2]{\normalcolor\checkitem #1 #2 \color{blue}}
\newcommand{\ifyespoints}[1]{\makebox[0pt][l]{\hspace{-15pt}\normalcolor #1}}

\end{document}
\fi
\end{document}